\documentclass[letterpaper, 10 pt, conference]{ieeeconf}  

\IEEEoverridecommandlockouts                              

\usepackage{amsmath} 
\usepackage{amssymb}  
\usepackage{ascmac}
\usepackage{xcolor}
\usepackage{graphicx}
\usepackage{cite}
\usepackage{algorithm}
\usepackage{algorithmic}
\usepackage{multirow}
\usepackage{subcaption}
\usepackage{url}
\usepackage{booktabs}
\usepackage{makecell}
\usepackage{hyperref}

\title{\LARGE \bf
DeReCo: Decoupling Representation and Coordination Learning \\ for Object-Adaptive Decentralized Multi-Robot Cooperative \\ Grasping and Lifting
}

\author{Kazuki Shibata$^{1*}$, Ryosuke Sota$^{1*}$, Shandil Dhiresh Bosch$^{1,2*}$,\\
Yuki Kadokawa$^{1}$, Yoshihisa Tsurumine$^{1}$, and Takamitsu Matsubara$^{1}$%
\thanks{$^{*}$These authors contributed equally to this work.}
\thanks{$^{1}$All the authors are with the Division of Information Science, Graduate School of Science and Technology, Nara Institute of Science and Technology (NAIST), Nara, Japan.}
\thanks{$^{2}$Shandil Dhiresh Bosch is with the Department of Cognitive Robotics, Faculty of Mechanical Engineering, Delft University of Technology, Delft, Netherlands.}
}

\begin{document}

\maketitle
\thispagestyle{empty}
\pagestyle{empty}

\begin{abstract}
Generalizing decentralized multi-robot cooperative grasping and lifting across objects with diverse shapes and physical properties remains a fundamental challenge. Under decentralized execution, two key challenges arise: object-dependent representation learning under partial observability and coordination learning in multi-agent reinforcement learning (MARL) under non-stationarity. A typical approach jointly optimizes object-dependent representations and coordinated policies in an end-to-end manner while randomizing object shapes and physical properties during training. However, this joint optimization tightly couples representation and coordination learning, which can introduce bidirectional interference: inaccurate representations under partial observability can degrade coordination learning, while non-stationarity in MARL can further degrade representation learning, leading to degraded task performance. Inspired by prior work on decoupling representation and control learning in single-robot settings, we propose DeReCo, a MARL framework that decouples representation and coordination learning for object-adaptive decentralized multi-robot cooperative grasping and lifting. DeReCo adopts a two-phase training strategy: (1) coordination learning with privileged object information, and (2) transition to decentralized execution through reconstruction of object-dependent representations from local observations and retraining of coordination policies using the reconstructed representations. Experimental results show that DeReCo outperforms baselines in simulation on three training objects, generalizes to six unseen objects with varying masses and friction coefficients under similar grasping configurations, and achieves superior performance on two unseen objects in real-robot experiments. The demonstration video is available at \href{https://sites.google.com/view/multi-hsrs-project/}{link}.
\end{abstract}

\section{INTRODUCTION}
This study considers decentralized multi-robot cooperative grasping and lifting, where mobile manipulators cooperatively grasp and lift an object to a target height using only local observations at execution time. These local observations exclude other robots' observations as well as privileged information about object-dependent properties such as shape, mass, and friction. Recent studies~\cite{Bernard2025,Feng2025,Pandit2024,Mehta2025,Zeng2025,Shibata2023RAS,Chen2025} adopt multi-agent reinforcement learning (MARL) to enable cooperative transport without relying on explicit dynamics models. However, most of these studies are evaluated in single-object settings and do not address generalization across objects with varying shapes and physical properties, particularly unseen objects not included during training. These limitations highlight the need for MARL approaches that generalize across diverse objects under decentralized execution.

\begin{figure}[t]
    \centering
    \includegraphics[width=0.95\columnwidth]{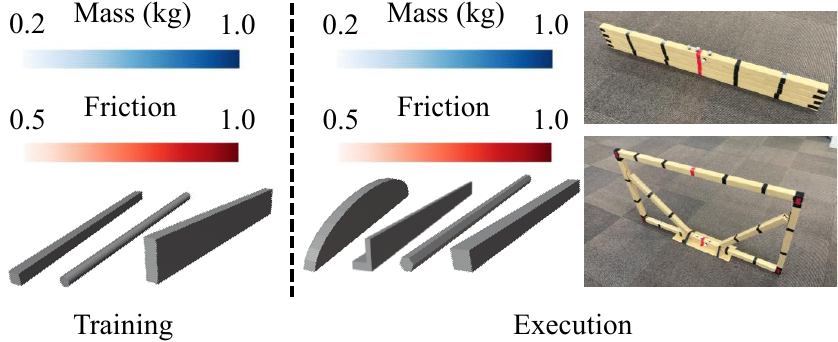}

    \smallskip
    {\footnotesize (a) Diverse objects\par}

    \vspace{0.5em}

    \includegraphics[width=0.95\columnwidth]{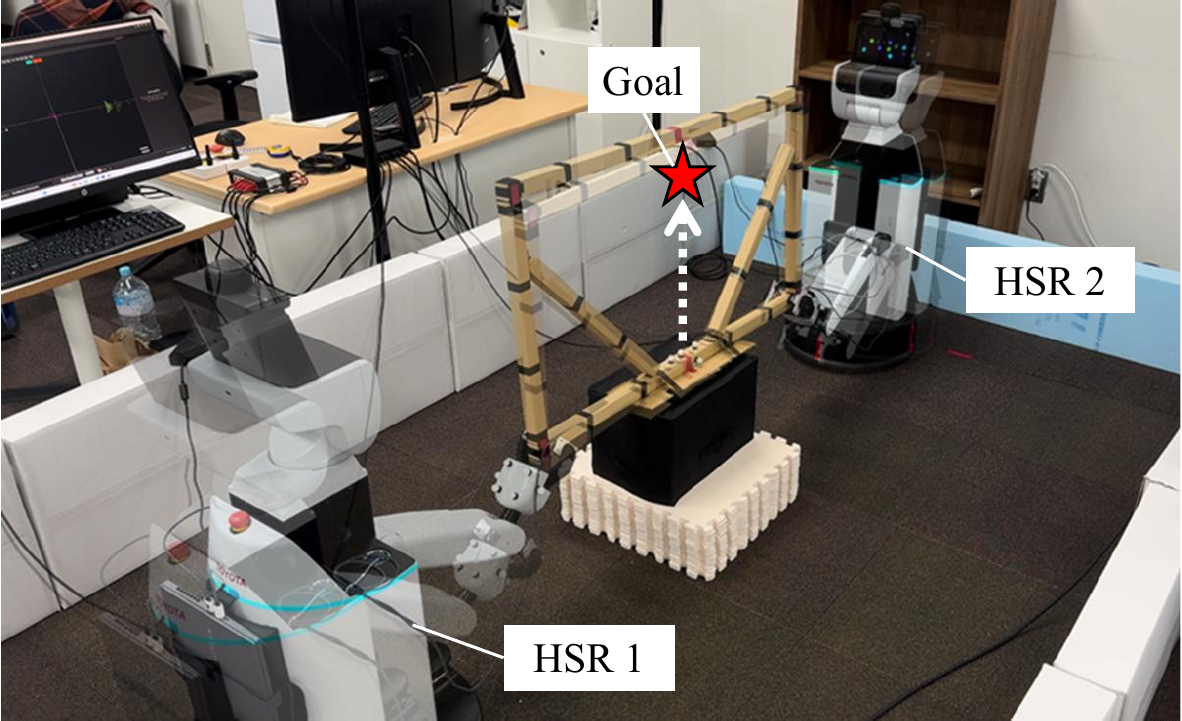}

    \smallskip
    {\footnotesize (b) Multi-robot cooperative grasping and lifting\par}

    \caption{\textbf{Multi-robot cooperative grasping and lifting of diverse objects using two HSRs.}}
    \label{fig:pull_figure}
\end{figure}

Generalizing decentralized multi-robot cooperative grasping and lifting across diverse objects poses two key challenges. First, object generalization under local observations constitutes a partial observability problem. If the environment state is defined to include object-dependent properties such as shape, mass, and friction, these properties are not directly observable at execution time. Robots must therefore infer object-dependent representations from limited sensory inputs. Second, cooperative grasping and lifting inherently requires coordination among multiple robots. In MARL, coordination learning is affected by non-stationarity because each agent's policy evolves during training, making the environment non-stationary from the perspective of the other agents~\cite{maddpg}.

A typical approach jointly optimizes object-dependent representations and coordinated policies in an end-to-end manner while randomizing object shapes and physical properties. This joint optimization tightly couples representation and coordination learning, which can introduce bidirectional interference: inaccurate representations under partial observability can degrade coordination learning, while MARL non-stationarity can further degrade representation learning, leading to degraded task performance.

To address this structural coupling, we decouple object-dependent representation learning under partial observability from coordination learning in MARL. Inspired by prior work on decoupling representation and control learning in single-robot settings, we propose DeReCo, a MARL framework that decouples representation and coordination learning for object-adaptive decentralized multi-robot cooperative grasping and lifting. DeReCo adopts a two-phase training strategy: (1) coordination learning with privileged object information, and (2) transition to decentralized execution through reconstruction of object-dependent representations from local observations and retraining of coordination policies using the reconstructed representations.

The effectiveness of DeReCo is confirmed through cooperative grasping and lifting experiments with two Human Support Robots (HSRs) in both simulation and the real world, as shown in Fig.~\ref{fig:pull_figure}. In simulation, we trained policies using three object shapes with different masses and friction coefficients and confirmed that DeReCo improves task performance compared to several baselines. We then evaluated generalization on six unseen object shapes that were not included during training, showing that the proposed method maintains high success rates on unseen objects with the same grasping-point configuration as the training objects. We further demonstrate superior performance over the strongest baseline in hardware experiments using two unseen objects that are longer than the training objects.

The main contributions of this work are as follows:
\begin{itemize}
\item We propose DeReCo, a MARL framework that decouples representation learning and coordination learning for object-adaptive decentralized multi-robot cooperative grasping and lifting.
\item We demonstrate through simulation that DeReCo consistently outperforms baseline methods across nine object shapes with varying masses and friction coefficients, including six shapes not included during training.
\item We evaluate the effectiveness of DeReCo on real HSRs and show superior performance over the strongest baseline on two unseen objects.
\end{itemize}

\section{Related Work}
\subsection{Model-based Multi-robot Cooperative Transport}
Model-based approaches to multi-robot cooperative transport assume explicit object dynamics and design controllers to drive the object toward a desired state. Representative methods include constrained optimal control~\cite{Hu2022,Koung2021,Kennel2024}, Model Predictive Control (MPC)~\cite{Li2023,Sihao2025}, consensus-based distributed optimization, distributed nonlinear MPC~\cite{Fawcett2023,Carli2025}, and decentralized adaptive control with Lyapunov-based stability guarantees~\cite{Culbertson2021,Yan2021,Sombolestan2023}.

These methods provide stability and convergence guarantees but generally assume rigid attachment between each robot and the object. In contrast, our model-free approach does not require an explicit dynamics model and remains effective without this rigid-contact assumption.

\subsection{MARL-based Multi-robot Cooperative Transport}
Several studies have demonstrated that model-free MARL enables cooperative transport across a wide range of robots, including mobile manipulators~\cite{Bernard2025}, quadrupeds~\cite{Feng2025}, bipeds~\cite{Pandit2024,Mehta2025}, and aerial platforms with cable-suspended payloads~\cite{Zeng2025}.

Most existing methods adopt centralized training with decentralized execution (CTDE), which uses global information during training to mitigate non-stationarity while executing decentralized policies from local observations. Scalability to varying team sizes has also been studied via consensus-based estimation~\cite{Shibata2023RAS} and curriculum learning with knowledge distillation~\cite{Chen2025}.

Despite these advances, most existing methods are evaluated in single-object settings and do not address generalization across diverse objects. In contrast, our approach decouples object-dependent representation learning under partial observability from coordination learning under MARL non-stationarity, enabling decentralized cooperative grasping and lifting across diverse objects.

\subsection{RL-based Transport of Diverse Objects}
Several studies have adopted RL-based approaches to achieve generalization in object transport. A typical approach employs domain randomization over object shapes and physical properties and jointly optimizes object-dependent representations and policy learning in an end-to-end manner~\cite{Peng2018,COCOI2021,Dadiotis2025}. Another line of work adopts a multi-stage learning framework that separates object-dependent representation learning from policy learning~\cite{Jeon2024}. Specifically, a policy is first trained with privileged object information; then an encoder is trained to reconstruct object-dependent representations from sensory inputs; finally, the policy is retrained using the reconstructed representations without privileged information.

Although prior work~\cite{Kumar2021,Jeon2024} separates representation learning from policy learning, these studies focus on single-robot learning and do not address decentralized multi-robot coordination. In contrast, our study decouples representation learning under partial observability from coordination learning in MARL under non-stationarity within the CTDE framework, enabling decentralized cooperative grasping and lifting across diverse objects.

\section{Preliminary}
\subsection{Problem Setting}
This study considers cooperative grasping and lifting with two mobile manipulators, where the robots grasp and lift objects with diverse shapes and physical properties. The control objective is to grasp an object and lift it to a specified target height.

Following~\cite{Bernard2025}, we make the following assumptions:
\begin{itemize}
  \item Each robot can observe the position and orientation of its end-effector and the object.
  \item The other robot’s grasping pose and contact forces are not available to each robot.
\end{itemize}

Moreover, we make the following assumption:
\begin{itemize}
  \item The object’s shape and physical properties are available during training, but are not accessible during execution.
\end{itemize}
This assumption is reasonable because this information is available in simulation during training, but is commonly unavailable in the real world.

\subsection{Decentralized Partially Observable Markov Decision Process (Dec-POMDP)}
In this study, we formulate the cooperative grasping and lifting problem as a Dec\mbox{-}POMDP~\cite{Dec-POMDP}. It is defined by the tuple
$\langle \mathcal{S}, \mathcal{B}, \{\mathcal{A}^i\}_{i\in\mathcal{B}}, P, \{r^i\}_{i\in\mathcal{B}}, \{\mathcal{O}^i\}_{i\in\mathcal{B}}, \gamma \rangle$,
where $s\in\mathcal{S}$ denotes the global state and $\mathcal{B}:=\{1,\dots,n\}$ is the set of robots.
Robot $i\in\mathcal{B}$ observes $o_t^i$ and selects $a_t^i$ at time $t$, and the actions of the $n$ robots form $a_t=(a_t^1,\dots,a_t^n)$. Given $(s_t,a_t)$, the next state $s_{t+1}$ is sampled from $P(s_{t+1}\mid s_t,a_t)$, and robot $i$ receives a reward $r_t^i$.

The learning objective is to maximize the expected discounted return across the $n$ robots, defined as $J=\mathbb{E}\!\left[\sum_{i=1}^{n} R_t^i\right]$, where $R_t^i=\sum_{k=0}^{\infty}\gamma^{k} r_{t+k}^i$ denotes the discounted return of robot $i$.

\subsection{Multi-Agent Proximal Policy Optimization (MAPPO)}
MAPPO~\cite{mappo} is a CTDE actor--critic algorithm. The actor of robot $i$ is parameterized by $\theta_i$ as $\pi_{\theta_i}(a_t^i \mid o_t^i)$, while the critic is parameterized by $\phi$ as $V_{\phi}(s_t)$.

Each actor is optimized using the clipped surrogate objective:
\begin{align}
\mathcal{L}_{\mathrm{actor}}(\theta_i)
= -\mathbb{E}_t\!\left[
\min\!\Big(
\rho_t^i A_t,\,
\mathrm{clip}(\rho_t^i,\,1-\epsilon,\,1+\epsilon)\,A_t
\Big)
\right],
\end{align}
where
$\rho_t^i=\pi_{\theta_i}(a_t^i|o_t^i)/\pi_{\theta_i^{\mathrm{old}}}(a_t^i|o_t^i)$
and $A_t$ is computed using generalized advantage estimation (GAE)~\cite{ppo}.

The critic is optimized with the clipped value loss:
\begin{align}
\mathcal{L}_{\mathrm{critic}}(\phi)
= \mathbb{E}_t \Big[
\max \Big(
(V_{\phi}(s_t)-\hat{R}_t)^2,\nonumber\\
(
\mathrm{clip}(
V_{\phi}(s_t),
V_{\phi}^{\mathrm{old}}(s_t)-\epsilon,
V_{\phi}^{\mathrm{old}}(s_t)+\epsilon
)
-\hat{R}_t
)^2
\Big)
\Big],
\end{align}
where $\hat{R}_t=\frac{1}{n}\sum_{i=1}^{n}R_t^i$ and $\epsilon$ is the clipping parameter.

\section{Method}
\subsection{Overview of DeReCo}
\begin{figure*}[t!]
    \centering
    \includegraphics[width=0.9\textwidth]{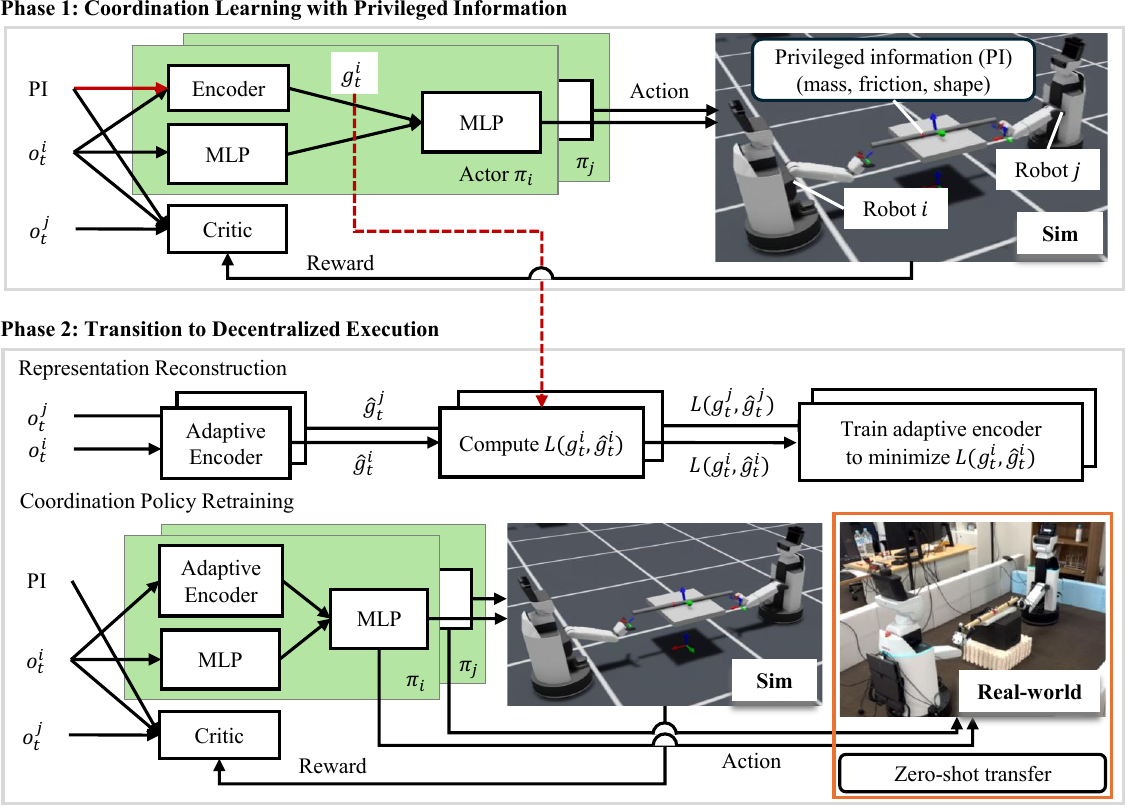}
    \caption{\centering Overview of the proposed framework for object-adaptive decentralized multi-robot cooperative grasping and lifting.}
    \label{fig:framework}
\end{figure*}

This section introduces DeReCo, a MARL framework for object-adaptive decentralized multi-robot cooperative grasping and lifting. Fig.~\ref{fig:framework} illustrates the overall structure of the proposed framework.

The training procedure of DeReCo consists of the following two phases:
\begin{itemize}
    \item \textbf{Phase 1: Coordination Learning with Privileged Information.}
    We learn coordination policies under centralized training using object-dependent representations from privileged object information.

    \item \textbf{Phase 2: Transition to Decentralized Execution.}
    We reconstruct object-dependent representations from local observations using an adaptive encoder and subsequently retrain the coordination policies under CTDE using the reconstructed representations, enabling decentralized execution without privileged information.
\end{itemize}

\subsection{Phase 1: Coordination Learning with Privileged Information}
We first perform centralized training with privileged information to learn coordination policies under MARL non-stationarity. This phase uses object-dependent representations without jointly learning representations from local observations.

During centralized training, the critic leverages joint observations and privileged object information $(o_t^i,\, o_t^j,\, p)$ to mitigate non-stationarity and is optimized by minimizing $\mathcal{L}_{\mathrm{critic}}$ to estimate the centralized value function. Here, $o_t^i$ and $o_t^j$ denote the local observations of robots $i$ and $j$ at time step $t$, and $p$ denotes privileged object information. The actor $\pi_\theta^i(a_t^i \mid o_t^i,\, p)$ is trained using the local observation and privileged information $(o_t^i,\, p)$ to learn coordination policies and is optimized by minimizing $\mathcal{L}_{\mathrm{actor}}$.

\subsection{Phase 2: Transition to Decentralized Execution}
Phase 2 transfers the coordination policy learned with privileged information in Phase~1 to decentralized execution through two complementary processes: representation reconstruction and coordination policy retraining. These processes form a unified transition to decentralized execution rather than independent performance-enhancing components.

\subsubsection{Representation Reconstruction}
Privileged object information used in centralized training is not available under decentralized execution. To decouple object-dependent representation learning under partial observability from coordination learning in MARL, the representation reconstruction process learns object-dependent representations from local observations via supervised learning. Specifically, this process consists of data collection and supervised training of the adaptive encoder.

\textbf{Data collection:}
We generate diverse objects by sampling discrete shape types from a predefined one-hot--encoded set and sampling continuous physical properties uniformly from specified ranges. For each sampled object, we roll out the policy from Phase~1 and record pairs $(o_t^i,\, g_t^i)$ for each robot $i$, where $g_t^i$ is the object-dependent representation computed during rollouts using the encoder in Phase~1, as shown in Fig.~\ref{fig:framework}. The representations $g_t^i$ serve as supervision targets for training the adaptive encoder.

\textbf{Supervised learning of the adaptive encoder:}
Under decentralized execution, each robot must infer object-dependent representations from its local observations without privileged object information. We therefore train an adaptive encoder $e_{\psi}$ that reconstructs object-dependent representations as $\hat{g}_t^i = e_{\psi}(o_t^i)$, where $\psi$ denotes the encoder parameters.

The adaptive encoder is trained in a supervised manner using target representations $g_t^i$ collected in Phase~1. We optimize $\psi$ by minimizing
$\mathcal{L}(g_t^i,\hat{g}_t^i)$, defined as the mean squared error between the reconstructed representations $\hat{g}_t^i$ and the targets $g_t^i$.

\subsubsection{Coordination Policy Retraining}
After reconstructing object-dependent representations from local observations, we retrain the coordination policy using the adaptive encoder to enable decentralized execution without privileged object information.

Under the CTDE framework, the actor and critic are initialized from the Phase~1 weights and retrained. Following~\cite{Jeon2024}, the adaptive encoder learned during representation reconstruction is frozen and used in the actor to reconstruct object-dependent representations from local observations. The critic leverages joint observations and privileged object information $(o_t^i,\, o_t^j,\, p)$ during training to mitigate non-stationarity and is optimized by minimizing $\mathcal{L}_{\mathrm{critic}}$. The actor $\pi_\theta^i(a_t^i \mid o_t^i)$ is trained using local observation $o_t^i$ to learn coordination policies and is optimized by minimizing $\mathcal{L}_{\mathrm{actor}}$.

At execution time, each robot selects actions using only its local observation. By reconstructing object-dependent representations from local observations and retraining the coordination policy using these representations, Phase~2 enables decentralized execution without privileged object information.

\section{Experiment}
To assess the effectiveness of DeReCo, we conduct a cooperative grasping and lifting experiment with two HSRs. This experiment aims to answer the following research questions:

\begin{itemize}
    \item \textbf{RQ1:} Can DeReCo improve training performance? (Section~\ref{sec:eval1})
    \item \textbf{RQ2:} Does DeReCo generalize to diverse objects with different shapes and physical properties? (Section~\ref{sec:eval2})
    \item \textbf{RQ3:} Can the trained policy successfully grasp and lift unseen objects in the real world? (Section~\ref{sec:demo})
\end{itemize}

\subsection{Experimental Setup}
\subsubsection{\bf{Simulation Setup}}
\label{sec:env_setting}
The simulations were conducted using Isaac Sim, an NVIDIA physics simulator. The control objective is for two HSRs to grasp and lift an object to a target height. Training was performed by running 512 parallel simulation environments in Isaac Sim.

The local observation of robot $i$ is defined as
$o^i = [a^i_{t-1},\, o^i_{\mathrm{arm}},\, o^i_{\mathrm{grip}},\, 
o^i_{\mathrm{goal}},\, o^i_{\mathrm{obj}},\, o^i_{\mathrm{force}}] \in \mathbb{R}^{27}$,
where $a^i_{t-1}\in \mathbb{R}^{6}$ is the previous action,
$o^i_{\mathrm{arm}}\in \mathbb{R}^{3}$ the arm joint angles,
$o^i_{\mathrm{grip}}\in \mathbb{R}^{2}$ the binary open/close states of the two gripper fingers,
$o^i_{\mathrm{goal}}\in \mathbb{R}^{3}$ the distance vector to the goal,
$o^i_{\mathrm{obj}}\in \mathbb{R}^{7}$ the object position and orientation relative to the end-effector, and
$o^i_{\mathrm{force}}\in \mathbb{R}^{6}$ the ternary force signals from two triaxial sensors on the gripper fingers, obtained by discretizing the force change between consecutive steps into $\{-1,0,+1\}$ following~\cite{Bernard2025}. 
The action is $a^i = [a^i_{\mathrm{base}},\, a^i_{\mathrm{arm}},\, a^i_{\mathrm{grip}}] \in \mathbb{R}^{6}$, 
where $a^i_{\mathrm{base}}\in \mathbb{R}^{2}$ and $a^i_{\mathrm{arm}}\in \mathbb{R}^{3}$ denote planar base and arm joint velocity commands, respectively, and $a^i_{\mathrm{grip}} \in \mathbb{R}$ applies a torque to the gripper fingers. 

The reward for robot $i$ ($i=1,2$) is designed following~\cite{Bernard2025}:
\begin{align}
r_i &= w_1 \cdot r_{\text{reach}} + w_2 \cdot r_{\text{grasp}} + w_3 \cdot r_{\text{grasp\_team}} \notag \\
    &\quad + w_4 \cdot r_{\text{track}} + w_5 \cdot r_{\text{ori}}
\end{align}
where $r_{\text{reach}}$ is a reward based on the distance between the end-effector and the grasp point, $r_{\text{grasp}}$ is obtained when each robot successfully grasps the object, $r_{\text{grasp\_team}}$ is given when both robots grasp the object simultaneously, $r_{\text{track}}$ is based on the distance between the current and target object positions, and $r_{\text{ori}}$ encourages the object to remain level with the ground during lifting. The weights $w_i$ ($i=1,\ldots,5$) are set to 3.0, 4.0, 7.5, 20, and 3.0, respectively.

At the beginning of each episode, the object is placed at the center of the table. The table height is sampled uniformly from $[0.3, 0.6]$\,m above the ground, and the goal position is fixed at a height of $0.8$\,m.

Furthermore, we evaluate performance using the success rate, where an episode is considered successful if the Euclidean distance between the object and the goal at the end of the episode is less than 0.1 m.

\subsubsection{\bf{Compared Methods}}
\label{sec:baseline}
To evaluate the proposed method, we compare it with the following MARL baselines:
\begin{itemize}
    \item \textbf{MAPPO w/o PI}: A MAPPO trained without privileged information (PI), using only local observations, in an end-to-end manner.

    \item \textbf{MAPPO w/o PI + LSTM}: A recurrent variant of \textbf{MAPPO w/o PI}, where an LSTM is integrated into the encoder to infer object-dependent representations.

    \item \textbf{MAPPO w/o AE}: A MAPPO trained without the adaptive encoder (AE) in an end-to-end manner. The critic is trained with privileged information, local observations, and other robots' observations, whereas the actor is trained only from local observations. 

    \item \textbf{MAPPO w/o AE + LSTM}: A recurrent variant of \textbf{MAPPO w/o AE}, where an LSTM is integrated into the encoder to infer object-dependent representations.

    \item \textbf{MAPPO w PI}: A MAPPO trained with both privileged information and local observations in an end-to-end manner.
\end{itemize}

\subsubsection{\bf{Network Architecture}}
The network architectures used in each phase of the proposed method are designed as follows.

\textbf{Phase 1: Coordination Learning with Privileged Information.}
The actor and critic share the same network architecture. Privileged information and local observations are processed by an encoder, which is a fully connected (FC) layer. Local observations are also passed through another FC layer. The outputs of the encoder and the FC layer are concatenated and then passed through two additional FC layers. Each FC layer has 128 units with ReLU activation in the hidden layers. The output layer uses a linear activation for the critic and a tanh activation for the actor.

\textbf{Phase 2: Transition to Decentralized Execution.}
Following \cite{Jeon2024}, the adaptive encoder is implemented as an LSTM. Specifically, we use a single-layer LSTM with 128 units.

The critic adopts the same network architecture as in Phase~1. The actor also follows the Phase~1 architecture, except that the encoder is replaced with the adaptive encoder. The actor and critic networks are initialized with the trained weights from Phase~1 and are then fine-tuned during Phase~2.

\subsection{RQ1: Can DeReCo Improve Training Performance?}
\label{sec:eval1}
\subsubsection{\bf Setup}
\begin{figure}[t]
    \centering
    \includegraphics[width=0.95\columnwidth]{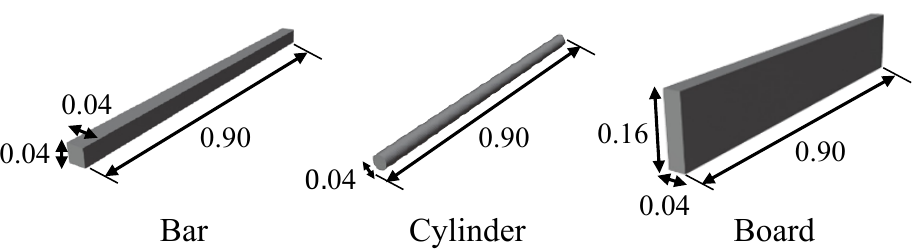}

    {\footnotesize (a) Seen objects\par}

    \vspace{0.5em}

    \includegraphics[width=0.95\columnwidth]{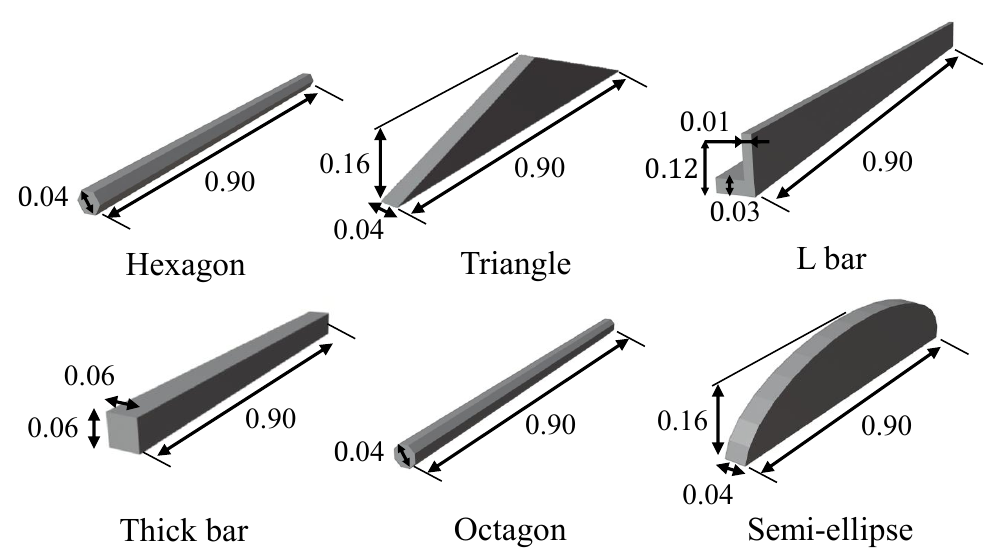}

    {\footnotesize (b) Unseen objects\par}

    \caption{\textbf{Objects used in the simulation. All units are meters.}}
    \label{fig:setup}
\end{figure}

To evaluate the effectiveness of DeReCo, we prepared a set of objects with different shapes and physical properties. During training, three object shapes were used, with masses ranging from 0.2~kg to 1.0~kg and friction coefficients ranging from 0.5 to 1.0. The objects used for training are shown in Fig.~\ref{fig:setup}(a). Training was performed for 50{,}000 steps in each phase of DeReCo and for 50{,}000 steps for each baseline method, where each step corresponded to one policy action. For both the proposed and compared methods, training was conducted five times with different random seeds, using an identical seed set across all methods to maintain consistency across runs.

\subsubsection{\bf Result}
\begin{figure}[t!]
    \centering
    \includegraphics[width=0.45\textwidth]{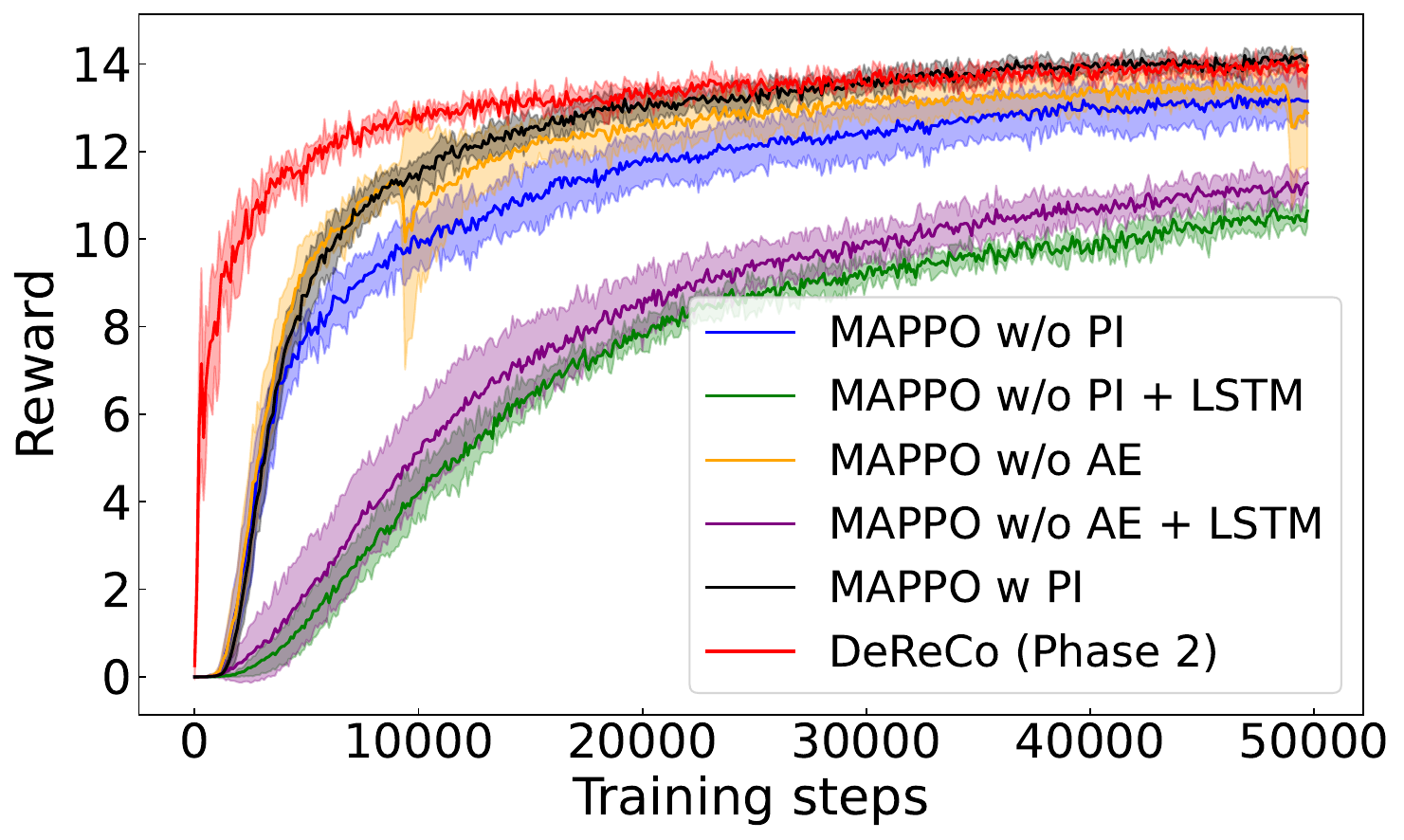}
    \caption{Comparison of tracking rewards. DeReCo (Phase 2) shows the learning curve during coordination policy retraining.}
    \label{fig:eval1_reward}
\end{figure}

Figure~\ref{fig:eval1_reward} compares the tracking reward during training. The curve labeled \textbf{DeReCo (Phase 2)} shows the tracking reward during coordination policy retraining. The learning curve for Phase~1 is omitted because its training procedure is identical to that of \textbf{MAPPO w PI}. \textbf{MAPPO w PI} achieves higher tracking rewards than \textbf{MAPPO w/o PI}, indicating the benefit of privileged object information for coordination learning. DeReCo achieves higher rewards than \textbf{MAPPO w/o AE}, supporting the effectiveness of the proposed two-phase training framework compared with end-to-end training. The recurrent baselines, \textbf{MAPPO w/o PI + LSTM} and \textbf{MAPPO w/o AE + LSTM}, exhibit lower rewards, suggesting that simply introducing recurrence into end-to-end training is insufficient to achieve comparable training performance.

Overall, DeReCo can improve training performance compared with the evaluated baselines.

\subsection{RQ2: Does DeReCo Generalize to Diverse Objects with Different Shapes and Physical Properties?}
\label{sec:eval2}
\subsubsection{\bf Setup}
We evaluate the policies on the three seen objects and six unseen objects with shapes not included in training, as shown in Fig.~\ref{fig:setup}(b). 
The object mass and friction coefficients are randomized within the same ranges as in training. All results are averaged over five independently trained policies, and each policy is evaluated for 1{,}000 trials using random seeds different from those used during training. We include \textbf{MAPPO w PI} as a reference, since it leverages privileged information at execution time. Because object IDs for unseen objects are not defined, we provide a random one-hot vector when evaluating \textbf{MAPPO w PI} on unseen objects.

\subsubsection{\bf Result}
Table~\ref{tab:seen_unseen_comparison} presents the per-object and average success rates on the seen and unseen objects. On the seen objects, \textbf{DeReCo} outperforms \textbf{MAPPO w/o PI} and achieves performance comparable to \textbf{MAPPO w PI} across objects with different shapes. This indicates that \textbf{DeReCo} achieves high coordination performance 
by inferring object-dependent representations from local observations.

\begin{table*}[t!]
\centering
\caption{Comparison of success rates on seen and unseen objects.
The symbol $\dagger$ indicates that privileged information (PI) was used for seen objects during execution.}
\label{tab:seen_unseen_comparison}
\begin{tabular}{lcccc ccccccc}
\toprule
\multirow{2}{*}{Method}
& \multicolumn{4}{c}{Seen Objects}
& \multicolumn{7}{c}{Unseen Objects} \\
\cmidrule(lr){2-5} \cmidrule(lr){6-12}
& Bar & Cylinder & Board & Avg.
& Hexagon & Triangle & L bar & Thick bar & Octagon & Semi-ellipse & Avg. \\
\midrule
MAPPO w/o PI
& 0.66 & 0.54 & 0.90 & 0.70
& 0.57 & 0.50 & 0.63 & 0.70 & 0.57 & 0.49 & 0.58 \\

MAPPO w/o PI + LSTM
& 0.55 & 0.26 & 0.46 & 0.42
& 0.33 & 0.39 & 0.48 & 0.21 & 0.30 & 0.44 & 0.36 \\

MAPPO w/o AE
& 0.81 & 0.69 & 0.95 & 0.82
& 0.72 & 0.61 & 0.74 & 0.72 & 0.71 & 0.60 & 0.68 \\

MAPPO w/o AE + LSTM
& 0.57 & 0.35 & 0.73 & 0.55
& 0.38 & 0.35 & 0.50 & 0.36 & 0.37 & 0.42 & 0.40 \\

MAPPO w PI$\dagger$
& 0.92 & \textbf{0.83} & \textbf{0.96} & \textbf{0.91}
& 0.44 & 0.31 & 0.53 & 0.60 & 0.47 & 0.33 & 0.45 \\

\textbf{DeReCo}
& \textbf{0.96} & \textbf{0.83} & 0.94 & \textbf{0.91}
& \textbf{0.85} & \textbf{0.70} & \textbf{0.86} & \textbf{0.85} & \textbf{0.86} & \textbf{0.69} & \textbf{0.80} \\
\bottomrule
\end{tabular}
\end{table*}

\begin{figure*}[t!]
    \centering

    \begin{minipage}[c]{0.035\textwidth}
        \centering
        \raisebox{1.3cm}{\rotatebox{90}{\small Failure Rate}}
    \end{minipage}%
    \begin{minipage}[c]{0.965\textwidth}
        \centering

        \includegraphics[width=0.6\linewidth]{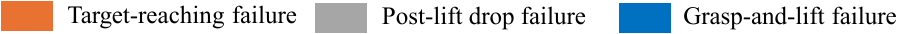}

        \vspace{0.5em}

        \begin{minipage}[t]{0.24\linewidth}
            \centering
            \includegraphics[width=\linewidth]{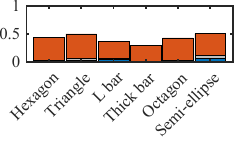}

            \smallskip
            {\footnotesize (a) \textbf{MAPPO w/o PI}\par}
        \end{minipage}%
        \hfill
        \begin{minipage}[t]{0.24\linewidth}
            \centering
            \includegraphics[width=\linewidth]{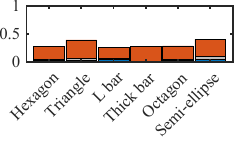}

            \smallskip
            {\footnotesize (b) \textbf{MAPPO w/o AE}\par}
        \end{minipage}%
        \hfill
        \begin{minipage}[t]{0.24\linewidth}
            \centering
            \includegraphics[width=\linewidth]{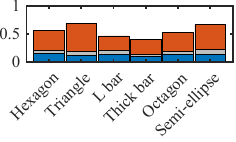}

            \smallskip
            {\footnotesize (c) \textbf{MAPPO w PI}\par}
        \end{minipage}%
        \hfill
        \begin{minipage}[t]{0.24\linewidth}
            \centering
            \includegraphics[width=\linewidth]{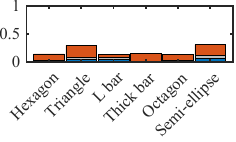}

            \smallskip
            {\footnotesize (d) \textbf{DeReCo}\par}
        \end{minipage}

    \end{minipage}

    \caption{\textbf{Failure analysis on unseen objects for four representative methods.}}
    \label{fig:failure}
\end{figure*}

On unseen objects, \textbf{DeReCo} consistently outperforms \textbf{MAPPO w/o PI}, demonstrating improved generalization to novel shapes. 
\textbf{DeReCo} also surpasses \textbf{MAPPO w/o AE}, suggesting that the proposed two-phase training framework improves generalization to unseen object shapes compared with end-to-end training. In contrast, \textbf{MAPPO w/o PI + LSTM} and \textbf{MAPPO w/o AE + LSTM} underperform \textbf{DeReCo}, indicating that jointly learning object representations and coordination end-to-end with LSTM remains insufficient; the proposed two-phase training framework achieves better generalization.

We further analyze failures on unseen objects for four representative methods as shown in Fig.~\ref{fig:failure}. Failures are categorized into three types: (i) grasping and lifting failures, where the robots cannot grasp and lift the object; (ii) post-lift drop failures, where the object is dropped after lifting; and (iii) target-reaching failures, where the object is lifted without dropping but does not reach within 0.1~m of the goal. The results show that target-reaching failures dominate across the six unseen objects, indicating that accurately reaching the goal is the primary challenge in this task. Notably, \textbf{MAPPO w PI} exhibits a significantly higher grasping and lifting failure rate than the other methods, likely because it is conditioned on privileged object IDs during training while unseen objects are assigned random IDs. This mismatch can induce inappropriate grasp strategies, suggesting that the poor performance of \textbf{MAPPO w PI} on unseen objects partly reflects this input mismatch.

Overall, the proposed method generalizes to diverse objects under decentralized execution by reconstructing object-dependent representations from local observations.

\subsection{RQ3: Can the Trained Policy Successfully Grasp and Lift Unseen Objects in the Real World?}
\label{sec:demo}

\subsubsection{\bf Setup}
\begin{figure}[t]
    \centering
    \begin{minipage}[t]{0.49\columnwidth}
        \centering
        \includegraphics[width=\linewidth]{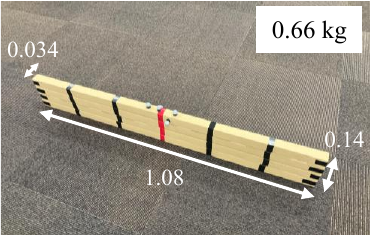}

        \smallskip
        {\footnotesize (a) Board\par}
    \end{minipage}%
    \hfill
    \begin{minipage}[t]{0.49\columnwidth}
        \centering
        \includegraphics[width=\linewidth]{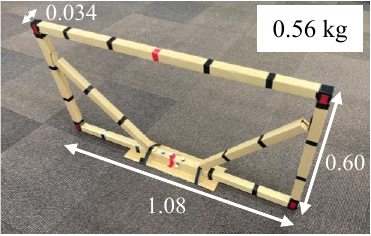}

        \smallskip
        {\footnotesize (b) Frame\par}
    \end{minipage}

    \caption{\textbf{Objects used in real experiments. All units are meters.}}
    \label{fig:demo-object}
\end{figure}

To confirm the effectiveness of DeReCo, we conducted hardware experiments with two HSRs and two unseen objects, as shown in Fig.~\ref{fig:demo-object}. The real-world objects were 1.08~m long, compared with 0.90~m for the training objects, allowing us to evaluate generalization to longer objects. We compared \textbf{DeReCo} with \textbf{MAPPO w/o AE}, as \textbf{MAPPO w PI} requires privileged object information that is unavailable during real-world execution. For each method, we deployed the best policy trained in simulation on the real robots. Each gripper finger was equipped with a GelSight Mini sensor at 60~Hz for contact-force measurement, and an OptiTrack motion-capture system at 120~Hz provided the poses of the robots’ end-effectors and the object. The observation and action were similar to those used in simulation. Success is defined as achieving a final object–goal distance error of less than 0.1 m. For fair comparison, the initial positions of the robots and the object were set identically across both methods.

\subsubsection{\bf Result}
\begin{figure*}[t]
    \centering

    \includegraphics[width=0.95\textwidth]{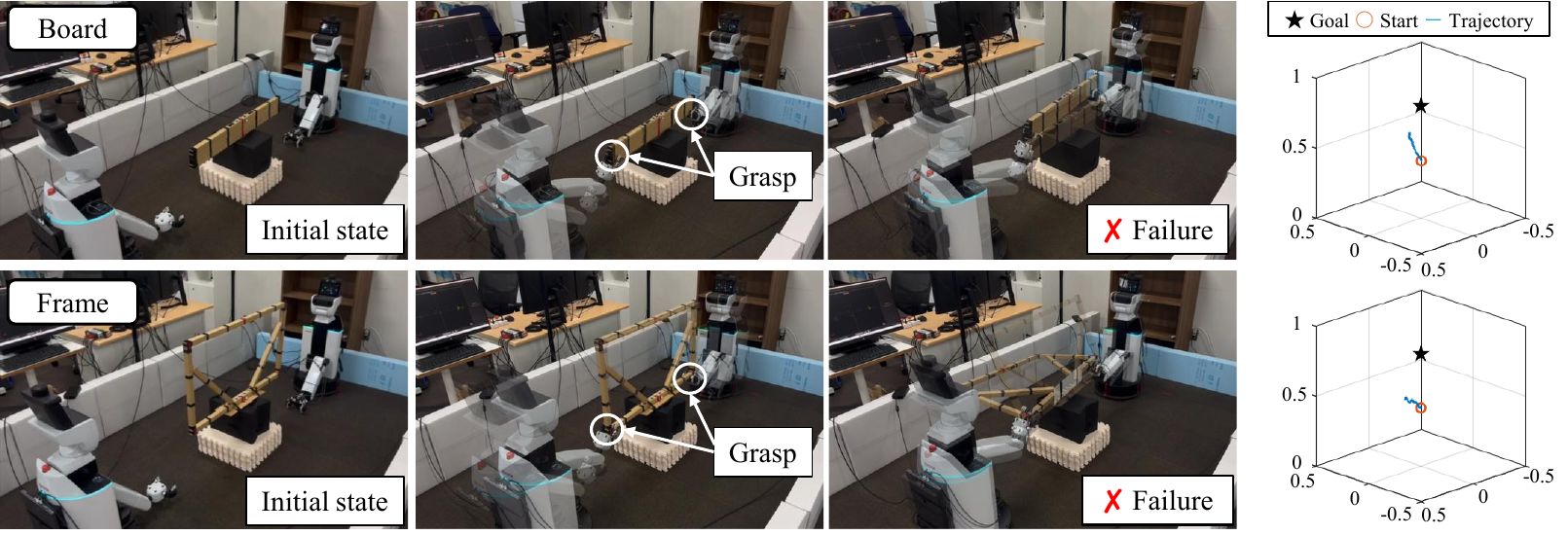}

    {\footnotesize (a) \textbf{MAPPO w/o AE}\par}

    \vspace{0.5em}

    \includegraphics[width=0.95\textwidth]{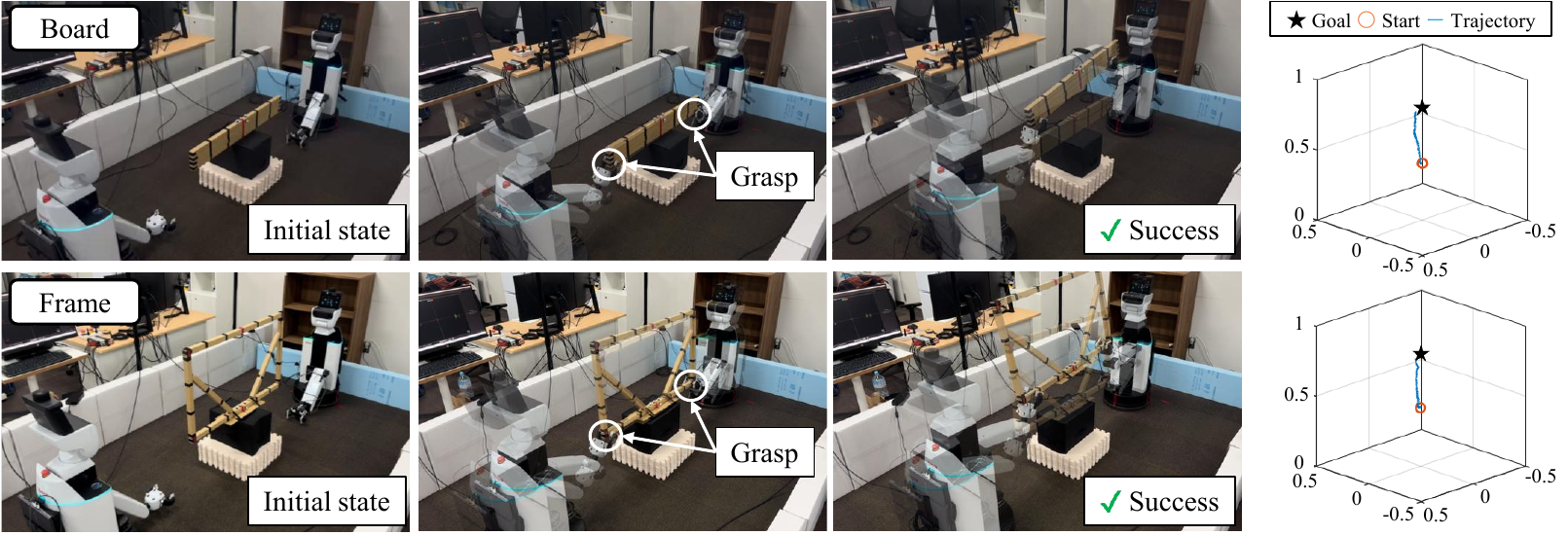}

    {\footnotesize (b) \textbf{DeReCo}\par}

    \caption{\textbf{Comparisons of real-robot demonstrations.}}
    \label{fig:demo}
\end{figure*}

Table~\ref{table:real-success} shows the success rates and final object--goal distance errors over five trials. Both \textbf{MAPPO w/o AE} and \textbf{DeReCo} achieved grasping and lifting for both objects in all five trials. However, \textbf{MAPPO w/o AE} failed to reach the target: it did not bring the board within 0.1~m of the goal, and the frame tipped over after lifting, as shown in Fig.~\ref{fig:demo}(a). In contrast, \textbf{DeReCo} successfully lifted both objects, as shown in Fig.~\ref{fig:demo}(b), and achieved final object--goal distance errors below 0.1~m on average for both objects. The single failure with the frame was a target-reaching failure, with a final object--goal distance error of 0.11~m, slightly exceeding the 0.1~m success threshold. Although the evaluation was limited to five trials per object, DeReCo demonstrated better performance than the baseline in these experiments.

Overall, the proposed method can successfully grasp and lift unseen objects in the real world.

\begin{table}[t]
    \centering
    \caption{Comparisons of real-robot experiments}
    \label{table:real-success}
    \begin{tabular}{l cc cc}
        \toprule
        \multirow{2}{*}{Method}
        & \multicolumn{2}{c}{Board} 
        & \multicolumn{2}{c}{Frame} \\
        \cmidrule(lr){2-3} \cmidrule(lr){4-5}
        & Success & Error [m]
        & Success & Error [m]  \\
        \midrule
        \textbf{MAPPO w/o AE} 
        & 0/5 & 0.18 $\pm$ 0.06
        & 0/5 & 0.34 $\pm$ 0.00 \\
        \textbf{DeReCo} 
        & \textbf{5/5} & \textbf{0.09} $\pm$ \textbf{0.01}
        & \textbf{4/5} & \textbf{0.09} $\pm$ \textbf{0.01} \\
        \bottomrule
    \end{tabular}
\end{table}

\section{Discussion}
This section introduces the limitations of DeReCo and outlines possible directions for future work.

We trained the policies on three representative shapes by randomizing the object mass and friction coefficients within predefined ranges. The proposed method achieves high success rates on unseen objects not included in training. However, the evaluated objects share the same grasping-point configuration, and generalization to objects requiring different contact configurations remains to be investigated. Furthermore, whether the adaptive encoder captures shape-related information beyond mass and friction has not been explicitly evaluated. In addition, extending the diversity of shapes and the ranges of mass and friction can degrade training performance, since extensive domain randomization increases training cost~\cite{Tiboni2024}. Scaling to a broader range of shapes and physical properties remains an important direction for future work.

Adapting to different numbers of robots is a crucial direction for future work. This study focuses on generalization across object shapes and physical properties in multi-robot cooperative grasping and lifting with a minimal two-robot setup. Our current MAPPO-based implementation uses a centralized critic whose input dimensionality depends on the number of robots, limiting its applicability to configurations with different team sizes.
 As future work, we will incorporate our approach into decentralized MARL formulations that do not depend on the number of robots, followed by real-robot validation of scalability~\cite{Shibata2023RAS,Chen2025}.

\section{Conclusion}
This paper proposed DeReCo, a MARL framework that decouples representation and coordination learning for object-adaptive multi-robot cooperative grasping and lifting. DeReCo adopts a two-phase training strategy: (1) coordination learning with privileged object information, and (2) transition to decentralized execution through reconstruction of object-dependent representations from local observations and retraining of coordination policies using the reconstructed representations. Simulation experiments with two HSRs across nine object shapes with varying masses and friction coefficients, including six unseen shapes, show that DeReCo outperforms baseline methods. Real-robot experiments on two unseen objects further confirmed successful sim-to-real transfer.

As future work, we plan to extend the proposed method to handle a wider range of object properties while quantitatively evaluating its performance in larger multi-robot teams.







\section*{Acknowledgments}
This work was supported by JSPS KAKENHI Grant Number JP25K21315 and the research grant program of The Futaba Foundation.

\bibliographystyle{IEEEtran}
\bibliography{reference}             

\end{document}